\documentclass[10pt,twocolumn,letterpaper]{article}

\usepackage{cvpr} 
\usepackage[accsupp]{axessibility}

\usepackage{microtype}
\renewcommand{\paragraph}[1]{\vspace{.5em}\noindent\textbf{#1.}} 

\usepackage{amssymb}
\usepackage{mathtools}
\usepackage{amsmath}
\usepackage{url}
\usepackage{booktabs}
\usepackage{graphicx}
\usepackage{caption}
\usepackage{tikz}
\usepackage{adjustbox}
\usepackage{enumitem}
\usepackage{comment}
\usepackage{textcomp}
\usepackage{xcolor}
\usepackage{subcaption} 
\usepackage{wrapfig}
\usepackage{makecell}

\usepackage{multirow}

\definecolor{myRed}{HTML}{CCCCCC} 
\definecolor{myGreen}{HTML}{00AA11}

\usepackage[framemethod=TikZ]{mdframed}
\definecolor{lightGray}{gray}{0.97}
\definecolor{midGray}{gray}{0.75}
\definecolor{lightYellow}{RGB}{254,254,239}
\definecolor{lighterYellow}{RGB}{254,254,245}
\definecolor{darkGray}{rgb}{0.45,0.45,0.45}
\definecolor{darkerGray}{rgb}{0.3,0.3,0.3}
\mdfdefinestyle{myFrameStyle}{%
  linecolor=midGray,
  linewidth=0.5pt,
  roundcorner=3pt,
  skipabove=5pt,
  skipbelow=0pt,
  innertopmargin=5pt,
  innerbottommargin=5pt,
  innerrightmargin=5pt,
  innerleftmargin=5pt,
  leftmargin=0pt,
  rightmargin=0pt,
  backgroundcolor=lighterYellow
}
\mdfdefinestyle{frameStyleQuestion}{%
  linecolor=midGray,
  linewidth=0.5pt,
  roundcorner=3pt,
  skipabove=5pt,
  skipbelow=0pt,
  innertopmargin=2.5pt,
  innerbottommargin=2.5pt,
  innerrightmargin=3pt,
  innerleftmargin=3pt,
  leftmargin=0pt,
  rightmargin=0pt,
  backgroundcolor=lightYellow
}

\newcommand{\myFrame}[1]{
    \vspace{4pt}
    \begin{mdframed}[style=myFrameStyle,userdefinedwidth=
    \linewidth,align=center,skipabove=6pt,skipbelow=0pt]
    {#1}
    \end{mdframed}\vspace{-8pt}
}

\renewcommand{\paragraph}[1]{\vspace{2pt}\noindent\textbf{#1}} 

\usepackage[hang,flushmargin]{footmisc} 

\newcommand{\liang}[1]{} 

\usepackage{siunitx} 
\usepackage[percent]{overpic}
\usepackage{pifont} 
\usepackage{array} 
\usepackage{makecell}

\makeatletter
\newcommand{\shorteq}{\mathrel{\mkern0.2mu\mathpalette\shorteq@\relax\mkern0.2mu}}
\newcommand{\shorteq@}[2]{\scalebox{0.5}[1]{$\m@th#1=$}}

\makeatother
\usepackage{bbold} 
\usepackage{nicefrac}
\newcommand\lldots{\ifmmode$\lldots$\else\thinspace\makebox[1em][c]{.\hfil.\hfil.}\fi} 

\definecolor{customblue}{HTML}{1A1AFF}
\definecolor{customred}{HTML}{FF4D4D}
\definecolor{darkYellow}{rgb}{0.8, 0.6, 0}

\makeatletter
\newcommand*{\rom}[1]{\expandafter\@slowromancap\romannumeral #1@}
\makeatother

\definecolor{cvprblue}{rgb}{0.21,0.49,0.74}
\usepackage[pagebackref,breaklinks,colorlinks,allcolors=cvprblue]{hyperref}

\def\paperID{42341} 
\def\confName{CVPR}
\def\confYear{2026}

\title{Can You Learn to See Without Images?\\Procedural Warm-Up for Vision Transformers}

\author{
Zachary Shinnick\thanks{Correspondence to: \texttt{zachary.shinnick@adelaide.edu.au}.}\, \textsuperscript{1} \quad
Liangze Jiang\textsuperscript{2,3} \quad
Hemanth Saratchandran\textsuperscript{1}\\
Damien Teney\textsuperscript{3} \quad
Anton van den Hengel\textsuperscript{1}\\[6pt]
\textsuperscript{1}Australian Institute for Machine Learning (AIML), University of Adelaide, Australia\\
\textsuperscript{2}École Polytechnique Fédérale de Lausanne (EPFL), Switzerland\\
\textsuperscript{3}Idiap Research Institute, Switzerland
}

\begin{document}
\maketitle
\begin{abstract}
Transformers are remarkably versatile,
suggesting the existence of generic inductive biases beneficial across modalities.
In this work, we explore a new way to instil such biases in vision transformers (ViTs)
through pretraining on procedurally generated data 
devoid of visual or semantic content.
We generate this data with simple algorithms such as formal grammars, so the results bear no relationship to either natural or synthetic images.
We use this procedurally generated data to pretrain ViTs in a warm-up phase that bypasses their visual patch embedding mechanisms,
thus encouraging the models to internalise abstract computational priors. 
When followed by standard image-based training, this warm-up significantly improves data efficiency, convergence speed, and downstream performance.
On ImageNet-1K, for example, allocating just 1\% of the training budget to procedural data improves final accuracy by over 1.7\%. 
In terms of 
its effect on performance, 1\% procedurally generated data is thus equivalent to 28\% of the ImageNet-1K data.
These findings suggest a promising path toward new data-efficient and domain-agnostic pretraining strategies. 

\vspace{3pt}
\noindent
{\small\textbf{Project page \& code:}
\url{https://zlshinnick.github.io/procedural-pretraining-page/}}\vspace{-9pt}
\end{abstract}

\section{Introduction}
\label{sec:intro}


Machine learning (ML) is moving towards increasingly generic approaches.
While early efforts in ML were often domain-specific,
recent work shows evidence of learning methods~\cite{goldblum2023no,teney2024neural},
computational mechanisms~\cite{jiang2026proceduralpretraining,shinnick2025transformers,han2025learning},
and representations~\cite{huh2024position}
that are applicable across tasks and modalities.
One prominent example is the transformer architecture~\cite{vaswani2017attention},
which implements generic inductive biases applicable to language as well as vision and other domains.
Besides architectural choices, acquiring domain-agnostic mechanisms \textit{from data}
could offer greater flexibility.
Yet this possibility remains underexplored.
Existing examples include training large language models (LLMs) on computer code
as a means to improve 
reasoning~\cite{petty2024does},
and the observation that
models trained on text acquire capabilities for handling visual data~\cite{han2025learning,sharma2024vision}. 
These examples suggest the existence of generic mechanisms independent from
any domain or modality.

\begin{figure}[t!]
    \vspace{1pt}
    \centering
    \includegraphics[width=1.0\linewidth]{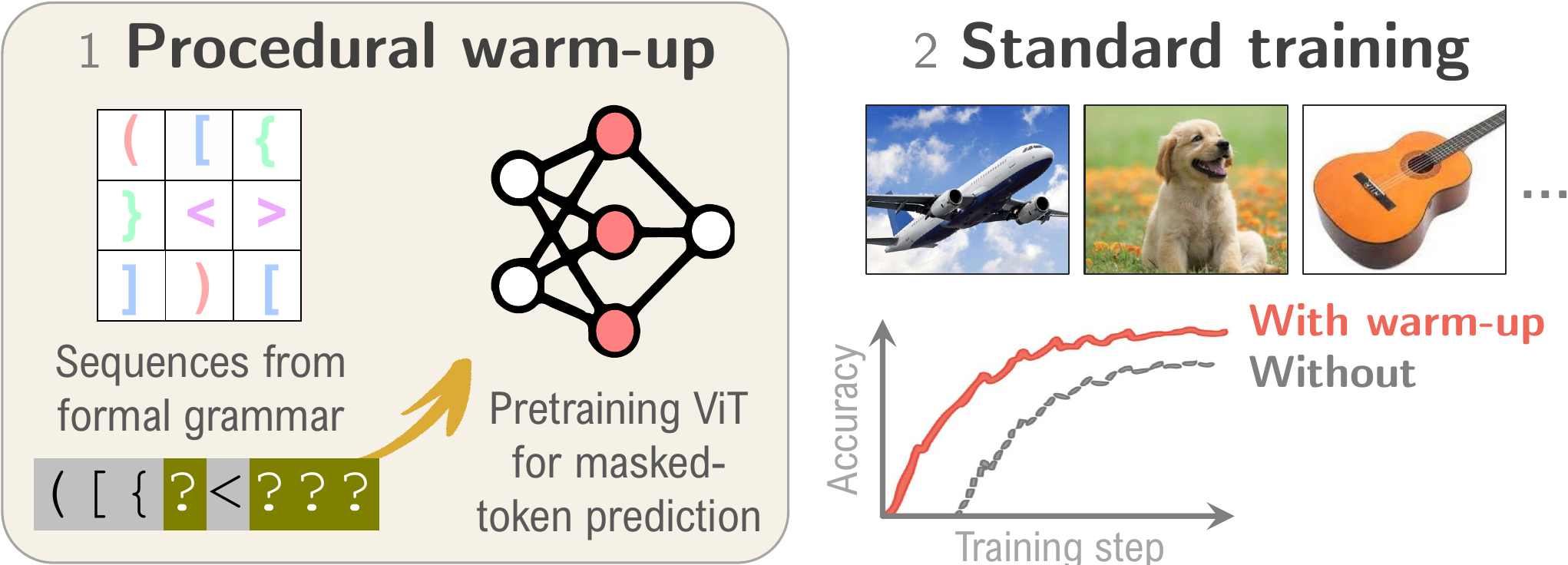}%
    \vspace{3pt}
    \caption{We propose a new pretraining phase for ViTs
    using synthetic data that is 
    symbolic and non-visual. 
    \textbf{(Left)} We generate procedural data
    consisting of sequences of abstract tokens using formal grammars
    (e.g.\ sequences of balanced parentheses).
    We bypass the ViT's visual patch embedding and
    directly train the model 
    for masked-token prediction.
    \textbf{(Right)}
    Followed by standard image-based training (e.g.\ on ImageNet), our
    lightweight warm-up consistently improves convergence and downstream performance.
    }
    \label{fig:teaser}
    \vspace{-10pt}
\end{figure}

\paragraph{Improving language models with abstract data.}
Recent work shows that language models can benefit from abstract data that \emph{imitates}
the structure of natural language, such as with procedural data
\footnote{\textit{Procedural data} refers to data generated with explicit algorithms, whereas \textit{synthetic data} typically refers to data from trained models,
e.g.\ text from LLMs~\cite{liu2024best,qin2025scaling} or images from GANs and diffusion models~\cite{rahimi2025auggen,azizi2023synthetic}.}
sampled from formal grammars~\citep{hu2025between}.
This data is devoid of any semantic information,
yet it can improve the convergence and performance 
of large language models (LLMs).
Other works show similar benefits with procedural data generated with other simple algorithms~\cite{jiang2026proceduralpretraining,wu2022insights,chen2024diversity} and cellular automata~\cite{zhang2024intelligence}
that can teach the model generic computational mechanisms.


\paragraph{Can vision models learn without images?}
We hypothesize that vision models could also benefit from procedural data,
as a means to acquire generic mechanisms relevant to vision tasks. 
However, it is unclear how vision models can `see' without images. 
Indeed, there is a rich literature on the training of vision models on
synthetic and/or abstract images \cite{rahimi2025auggen,sixt2018rendergan,ho2020denoising,tremblay2018training,richter2016playing}.
These works typically rely on handcrafted algorithms that generate abstract images
imitating properties of natural data~\cite{ruderman1997origins}.
These images have mostly been evaluated as a replacement for standard datasets
to address privacy and fairness concerns.
In this paper, we go a step further and completely forgo the constraints of visual data.
Instead, we train vision models on procedural data with no 2D structure or explicit correspondence with image properties. The hope is that this data might capture even more generally useful structure as a result. The problem of reasoning over images is primarily a reasoning problem, not an image problem.


\paragraph{Proposed method and results.}
We propose a training pipeline for vision transformers (ViTs)
that leverages procedural data in a brief warm-up prior to standard training on natural images (see Figure~\ref{fig:teaser}).
Our procedural data is generated with simple formal grammars, contains no semantic information,
and does not attempt to imitate visual structures.
This data is cheap to generate and process. To feed it to ViTs, we 
bypass the visual patch embeddings, and map instead the abstract symbols
to random frozen embeddings, which are later discarded (see Section~\ref{sec:methods}).
We conduct an extensive evaluation of the value of this abstract data for
image classification tasks.
We generally find substantial improvements in downstream performance.
Allocating as little as 1\% of the total training budget to procedural data 
yields a +1.7\% improvement in accuracy on \textsc{ImageNet-1k}.
This means that it can replace up to 28\% of the image data for equivalent performance.
The improvements also persist when the model is fine-tuned on various
domain-specific datasets.
This suggests that the information learned from procedural data
is distinct and complementary to standard visual learning,
rather than an analogue or substitute.

\paragraph{Detailed analysis.}
With careful ablations, we verify that the benefits arise from structured dependencies within the procedural data.
We also find that procedural warm-up acts both on attention and MLP layers,
contrasting with existing structured initializations of transformers, which act only on the attention weights~\cite{trockman2023mimetic,zheng2025structured, giri2025ibitutilizinginductivebiases}.
Procedural warm-up also acts mainly on the \emph{latter} layers, which contrasts with the established knowledge that \emph{early} layers are most important for standard visual pretraining~\cite{lenc2015understanding}.
This supports the conclusion that procedural warm-up provides a qualitatively different
type of training signal than standard data.


\vspace{2pt} 
\noindent
Our contributions are summarized as follows.
\begin{itemize}[leftmargin=*, itemsep=2pt, parsep=0pt, topsep=1pt]
\item \textbf{Procedural warm-up for vision transformers.}
We propose a lightweight pretraining stage for ViTs using symbolic, non-visual data generated with formal grammars, as a means to acquire generic mechanisms that support subsequent visual learning.
This expands recent work on procedural pretraining~\cite{jiang2026proceduralpretraining, hu2025between} from LLMs to ViTs.

\item
\textbf{Empirical evaluation.}
We implement our approach with standard ViTs and evaluate it on various benchmarks.
We show that a brief procedural warm-up consistently accelerates convergence and improves the accuracy of the models.
The gains persist when combined with large-scale pretraining on ImageNet,
suggesting that procedural warm-up and visual pretraining provide complementary training signals.

\item
\textbf{Analysis of the source of improvements.}
We show that the benefits arise from the precise structure of the procedural data, and manifest in both attention and MLP layers, particularly in later layers. This surprising observation contrasts with standard pretraining which acts primarily on \emph{early} layers. This suggests that procedural warm-up is qualitatively different
from standard visual pretraining or existing structured initialization methods.
\end{itemize}

\section{Related Work}
\label{sec:related_work}

\begin{table*}[t]
    \centering
    \small
    \setlength{\tabcolsep}{10pt}
    \renewcommand{\arraystretch}{1.0}
    \caption{\textbf{Formal languages used to generate our procedural data.}
    We select standard simple grammars representative of different levels of structural complexity in the Chomsky hierarchy, from regular (\textsc{WW}) to context-free (\textsc{$k$-Dyck}) and context-sensitive (\textsc{$k$-Dyck Shuffle}).}
    \begin{tabular}{l c >{\centering\arraybackslash}m{0.25\textwidth} l}
    \toprule
    \textbf{Language} & \textbf{Type} & \textbf{Example sequence} & \textbf{Structural property} \\
    \midrule
    
    \textsc{WW} & Regular &
    \texttt{%
    \textbf{%
    \textcolor{blue}{a b c }%
    \textcolor{red}{a b c}%
    }%
    }
    & Global one-to-one correspondence \\
    
    \textsc{$k$-Dyck} & Context-free & 
    \texttt{%
    \textcolor{blue}{\textbf{( }}%
    \textcolor{red}{\textbf{[ }}%
    \textcolor{red}{\textbf{] }}%
    \textcolor{blue}{\textbf{) }}%
    \textcolor{black}{\textbf{< }}%
    \textcolor{black}{\textbf{>}}%
    }
    & Hierarchical, stack-based dependencies \\
    
    \textsc{$k$-Dyck Shuffle} & Context-sensitive &
    \texttt{%
    \textcolor{blue}{\textbf{( }}%
    \textcolor{red}{\textbf{[ }}%
    \textcolor{blue}{\textbf{) }}%
    \textcolor{black}{\textbf{< }}%
    \textcolor{red}{\textbf{] }}%
    \textcolor{black}{\textbf{>}}%
    }
    & Crossing and interleaved dependencies \\
    
    \bottomrule
    \end{tabular}
    \label{tab:formal_languages}
    \vspace{-4pt}
    \end{table*}

\paragraph{Training vision models with abstract images.}
The properties of natural images have been extensively studied~\cite{ruderman1997origins}.
Many procedural algorithms exist to generate images that
imitate structural and statistical properties of natural data~\cite{baradad2022procedural}.
Abstract images used to train vision models include
fractals~\cite{nakamura2023pretrainingvisiontransformerslimited,nakamura2024scaling},
contours~\cite{kataoka2022replacinglabeledrealimagedatasets},
wave patterns~\cite{takashima2023visualatomspretrainingvision},
and structured noise~\cite{baradad2022learninglookingnoise}.
The motivation 
is usually to replace natural images
for privacy and fairness reasons~\cite{kataoka2021pretrainingnaturalimages},
or to allow learning from smaller datasets~\cite{nakashima2023formuladriven}.
Our work differs because we use \emph{non-visual} data,
and our goal is to \emph{complement} natural images by providing a different
training signal.


\paragraph{Vision capabilities from non-visual data.}
Recent work shows that some text-only models are capable of
processing visual information~\cite{du2025large,zheng2024lm4lv}, sometimes after minimal adaptation~\cite{li2023blip,tong2025metamorph}.
\citeauthor{han2025learning} show that these effects arise from
the general knowledge present in text data (e.g.\ descriptions of visual attributes)
but also from the transfer of modality-agnostic reasoning mechanisms.
This motivates our attempt to improve vision capabilities from non-visual data.

\paragraph{Initialization of vision models.}
The initial weights of neural networks are critical for the convergence of gradient-based training~\cite{huang2020improving}.
Besides simple adjustments to the distribution 
of random weights,
structured initializations exist
to improve the training speed or generalization of transformers~\cite{huang2020improving,zhao2022zero,trockman2023mimetic}.
The \textit{Mimetic} initialization 
creates diagonal patterns in attention matrices
that imitate 
trained models.
Other methods initialize the attention of ViTs to imitate the behaviour of CNNs~\cite{zheng2025structured,giri2025ibitutilizinginductivebiases}.
These approaches affect only the attention layers, while our method acts on the entire network.
They all aim to instantiate structures 
presumed to be useful for subsequent vision tasks.

\paragraph{Procedural data in other domains.}
The NLP literature (natural language processing)
is rich in attempts to train models with abstract synthetic data
e.g.\ for machine translation~\citep{he2023synthetic},
mathematical reasoning~\cite{wu2021lime},
text encoding~\cite{ri2022pretraining},
and general-purpose language modeling.
Formal grammars are a common approach to sample data devoid of semantic meaning
that imitates properties of natural language
\citep{chiang2022transferability,goodale2025meta,mccoy2023modeling,papadimitriou2023injecting,ri2022pretraining,hu2025between}.
Recent work also uses data generated with other simple algorithms including cellular automata
\citep{lindemann2024sip,wu2022insights,wu2021lime,zhang2024intelligence}. 
Most similar to our work,
\cite{jiang2026proceduralpretraining} and \cite{shinnick2025transformers}
propose a procedural warm-up stage for LLMs, and find that it provides training information
complementary to standard datasets,
which yields improved performance
with natural language, computer code, and informal mathematics.
Finally, procedural data has also been used to train models for
visual navigation~\citep{wang2022visual},
and reinforcement learning~\citep{baradad2022procedural}.
\section{Proposed Method}
\label{sec:methods}
This section presents our method for procedural warm-up of ViTs.
First, we generate procedural data with formal languages (Section~\ref{sec:data_generation}).
This data is symbolic, non-visual, and easily predictable given the knowledge of the generating grammar.
Second, we use this data in a \textit{lightweight} pretraining stage
where a ViT is trained for standard masked-token prediction (Section~\ref{sec:pretraining}).
This drives the model to discover the data-generating algorithm
and thereby learn generic computational mechanisms.
We will show in Section~\ref{sec:experiments}
that the resulting model is a much better starting point for standard image-based training
than a typical random initialization, improving both its training speed and final performance.

\subsection{Generating Procedural Data}
\label{sec:data_generation}

Our goal is to cheaply generate data that contains predictable structures with different levels of complexity.
Formal languages offer a simple specification of a flexible set of data-generating algorithms.
Moreover, data generated with formal languages has been used with success in LLMs~\cite{hu2025between,jiang2026proceduralpretraining} as a complement to natural language.

\paragraph{Selected formal grammars.}
For our study, we select a set of simple grammars summarized in Table~\ref{tab:formal_languages}.
They span multiple levels of complexity in the \textit{Chomsky hierarchy}~\cite{chomsky1959formal}. The levels of this hierarchy correspond to the computational machinery necessary to recognise these languages: \textit{regular} languages can be processed by finite-state automata, \textit{context-free} languages require a stack to capture hierarchical structure,
and \textit{context-sensitive} languages extend this further to represent cross-serial and multi-level dependencies.
Consequently, data generated with our chosen grammars contain a variety of structures,
from simple regular repetition to nested and cross-serial dependencies:

\begin{itemize}[leftmargin=*]
\item \textbf{\textsc{WW}.} A simple \textit{regular} language consisting of a string concatenated with its exact copy.

\item \textbf{\textsc{$k$-Dyck}.} A \textit{context-free} language that generates balanced parentheses with a hierarchical structure. The parameter $k$ controls the number of unique bracket pairs, which determines the variety and complexity of the nesting.

\item \textbf{\textsc{$k$-Dyck Shuffle}.} Relaxes the \textsc{$k$-Dyck} grammar by removing the constraint that brackets must be well-nested, while preserving the requirement that every opening bracket is properly closed. This yields a \textit{context-sensitive} language with crossing dependencies, representing multiple intertwined hierarchies.
\end{itemize}

\paragraph{Generating token sequences for ViTs.}
For any chosen grammar, we generate sequences using a stochastic generator matched to its structure: \textsc{WW} sequences are formed by sampling a random substring and concatenating it with an exact copy;
\textsc{$k$-Dyck} strings are produced with a stack-based process that randomly opens and closes brackets while enforcing proper nesting;
and \textsc{$k$-Dyck Shuffle} sequences are generated by stochastically opening new brackets or closing currently open ones. We use a vocabulary of 128 tokens (unless stated otherwise), following \cite{hu2025between}, which means setting $k\!=\!64$ for the Dyck languages.
ViTs are designed to take in a fixed number of visual tokens,
thus, each procedural sequence is generated with a fixed size of
$N\!\!=\!\!H\!\times\!W$.
Importantly, this symbolic data can be generated 
at a {negligible computational cost}, enabling a lightweight warm-up stage
with no natural images.

\subsection{Pretraining ViTs with Procedural Data}
\label{sec:pretraining}

\paragraph{Feeding abstract tokens to ViTs.}
Our study uses standard ViTs~\cite{dosovitskiy2020image},
which start with a linear projection known as the patch embedding layer.
This normally projects the image contents to a sequence of feature vectors.
Since our procedural data is made of tokens from a discrete vocabulary,
we use instead a discrete embedding layer as found in language models (i.e.\ a lookup table).
We keep this embedding fixed with random vectors,
such that the symbols from the vocabulary are assigned to approximately orthogonal vectors.
This prevents the model from solving the pretraining objective through the embeddings,
forcing it instead to adapt its attention and MLP layers.
We similarly maintain the positional encoding frozen during procedural pretraining.

\paragraph{Pretraining objective.}
The procedural warm-up uses a standard masked-token objective with a token prediction head. For each sequence, we mask 50\% of structurally informed tokens, such as closing brackets in \textsc{Dyck} and \textsc{Dyck Shuffle}, or the repeated sequence in \textsc{WW}, and require the model to predict the original token.
As mentioned above, we keep the embedding layer frozen during the procedural pretraining and only update the attention and MLP layers.

\paragraph{Standard image-based training.}
After the procedural warm-up,
we proceed with standard pretraining and/or fine-tuning on natural images.
We discard the token embeddings and prediction head that were necessary for the procedural data.
We then update the entire architecture using standard learning rates and hyperparameters.
The procedurally-pretrained model can thus be compared to other forms of weight initialization. 

\paragraph{Note on ambiguous grammars.}
For the \textsc{$k$-Dyck} and \textsc{WW} languages, we use a loss mask
such that each example has a unique valid completion
(masking opening parentheses and the first half of sequences, respectively).
For \textsc{$k$-Dyck Shuffle}, each sequence admits multiple valid completions.
For example, `\texttt{( [ Mask < Mask >}' could be completed with `\texttt{( [ ] < ) >}' or `\texttt{( [ ) < ] >}'.
We train the model with ``teacher-forcing'' supervision on one valid continuation for each training example.
When trained on this data, the model thus cannot reach perfect accuracy,
yet it drives the model to learn to track stack depth and bracket types 
to assign non-zero probability to valid continuations.
Overall, learning valid completions for our formal languages
requires learning and implementing 
generic mechanisms for hierarchical composition, long-range dependencies, and stack manipulation. 


\paragraph{Summary.}
Our procedural warm-up (i)~uses procedurally generated symbolic data from formal languages,
(ii)~freezes the embedding layers, ensuring that the model encodes new knowledge
in its attention and MLP layers, and
(iii)~uses the pretrained weights as initialization for standard image-based training.
The setup allows us to acquire knowledge from abstract symbolic data
and leverage it for vision tasks.
The next section empirically evaluates the value of this abstract knowledge on a variety of use cases and datasets.


\section{Experiments}
\label{sec:experiments}
\begin{table*}[t]
\centering
\small
\setlength{\tabcolsep}{6pt}
\renewcommand{\arraystretch}{1.0}
\caption{\textbf{Accuracy (\%) on image-classification benchmarks.}  
Models subject to procedural warm-up consistently converge to a higher accuracy
than other generic initialisation/pretraining schemes.
Green subscripts denote absolute improvements over the default initialization.}
\begin{tabular}{lcccccc}
\toprule
\textbf{Method} & \textbf{\textsc{ImageNet-1k}} & \textbf{\textsc{Tiny-ImageNet}} & \textbf{\textsc{Food-101}} & \textbf{\textsc{CIFAR-10}} & \textbf{\textsc{CIFAR-100}} & \textbf{\textsc{STL-10}} \\
\midrule
Default random initialization~\cite{dosovitskiy2020image} &
77.49\textsubscript{\textcolor{white}{\footnotesize +0.00}} &
55.42\textsubscript{\textcolor{white}{\footnotesize +0.00}} &
74.52\textsubscript{\textcolor{white}{\footnotesize +0.00}} &
91.29\textsubscript{\textcolor{white}{\footnotesize +0.00}} &
68.52\textsubscript{\textcolor{white}{\footnotesize +0.00}} &
60.52\textsubscript{\textcolor{white}{\footnotesize +0.00}} \\

Mimetic initialization~\cite{trockman2023mimetic} &
78.68\textsubscript{\textcolor{myGreen}{\footnotesize\,+1.19}} &
57.20\textsubscript{\textcolor{myGreen}{\footnotesize\,+1.78}} &
79.21\textsubscript{\textcolor{myGreen}{\footnotesize\,+4.69}} &
\textbf{92.89\textsubscript{\textcolor{myGreen}{\footnotesize\,+1.60}}} &
70.72\textsubscript{\textcolor{myGreen}{\footnotesize\,+2.20}} &
65.37\textsubscript{\textcolor{myGreen}{\footnotesize\,+4.85}} \\

FractalDB warm-up~\cite{kataoka2021pretrainingnaturalimages}&
78.06\textsubscript{\textcolor{myGreen}{\footnotesize\,+0.57}} &
55.17\textsubscript{\textcolor{myRed}{\footnotesize\, -0.25}} &
74.25\textsubscript{\textcolor{myRed}{\footnotesize\, -0.27}} &
88.98\textsubscript{\textcolor{myRed}{\footnotesize\, -2.31}} &
64.61\textsubscript{\textcolor{myRed}{\footnotesize\, -3.91}} &
58.62\textsubscript{\textcolor{myRed}{\footnotesize\, -1.90}} \\

\midrule
\textbf{Procedural warm-up (ours)}&
\textbf{79.21\textsubscript{\textcolor{myGreen}{\footnotesize\,+1.72}}} &
\textbf{58.20\textsubscript{\textcolor{myGreen}{\footnotesize\,+2.78}}} &
\textbf{79.47\textsubscript{\textcolor{myGreen}{\footnotesize\,+4.95}}} &
\underline{92.81}\textsubscript{\textcolor{myGreen}{\footnotesize\,+1.52}} &
\textbf{71.98\textsubscript{\textcolor{myGreen}{\footnotesize\,+3.46}}} &
\textbf{66.48\textsubscript{\textcolor{myGreen}{\footnotesize\,+5.96}}} \\
\bottomrule
\end{tabular}
\label{tab:symbolic-downstream}
\end{table*}

\begin{table*}[t]
\caption{\textbf{Accuracy (\%) of ViT-B models pretrained on \textsc{ImageNet-1k} and fine-tuned on various datasets.}  
The procedural warm-up consistently improves the transfer performance compared to the default, Mimetic and FractalDB baselines.
Green subscripts denote absolute improvements over the default ImageNet initialization.
The fact that the improvements persist throughout large-scale pretraining and fine-tuning stages suggests that the procedural data provides a training signal distinct and complementary to natural images.}
\centering
\small
\setlength{\tabcolsep}{6.0pt}
\renewcommand{\arraystretch}{1.0}
\begin{tabular}{lccccc}
\toprule
\textbf{Method} & \textbf{\textsc{Tiny-ImageNet}} & \textbf{\textsc{Food-101}} & \textbf{\textsc{CIFAR-10}} & \textbf{\textsc{CIFAR-100}} & \textbf{\textsc{STL-10}} \\
\midrule
Random init.\ \cite{dosovitskiy2020image} + \textsc{ImageNet-1k} & 86.59\textsubscript{\textcolor{white}{\footnotesize +0.00}} & 89.64\textsubscript{\textcolor{white}{\footnotesize +0.00}} & 98.59\textsubscript{\textcolor{white}{\footnotesize +0.00}} & 87.54\textsubscript{\textcolor{white}{\footnotesize +0.00}}& 98.55\textsubscript{\textcolor{white}{\footnotesize +0.00}} \\
Mimetic init.\ \cite{trockman2023mimetic} + \textsc{ImageNet-1k} & 87.29\textsubscript{\textcolor{myGreen}{\footnotesize +0.70}} & 90.74\textsubscript{\textcolor{myGreen}{\footnotesize +1.10}} & \textbf{98.68}\textsubscript{\textcolor{myGreen}{\footnotesize \textbf{+0.09}}} & 88.78\textsubscript{\textcolor{myGreen}{\footnotesize +1.24}} & \textbf{98.81}\textsubscript{\textcolor{myGreen}{\footnotesize \textbf{+0.26}}} \\
FractalDB~\cite{kataoka2021pretrainingnaturalimages} + \textsc{ImageNet-1k} & \textbf{88.42}\textsubscript{\textcolor{myGreen}{\footnotesize \textbf{+1.83}}} & 90.13\textsubscript{\textcolor{myGreen}{\footnotesize +0.49}} & 98.41\textsubscript{\textcolor{myRed}{\footnotesize -0.18}} & 88.35\textsubscript{\textcolor{myGreen}{\footnotesize +0.81}} & 98.46\textsubscript{\textcolor{myRed}{\footnotesize -0.09}} \\
\midrule
\textbf{Procedural warm-up (ours)} + \textsc{ImageNet-1k} & \underline{87.93}\textsubscript{\textcolor{myGreen}{\footnotesize +1.34}} & 
\textbf{90.79}\textsubscript{\textcolor{myGreen}{\footnotesize \textbf{+1.15}}} & 
\textbf{98.68}\textsubscript{\textcolor{myGreen}{\footnotesize \textbf{+0.09}}} & 
\textbf{89.20}\textsubscript{\textcolor{myGreen}{\footnotesize \textbf{+1.66}}} & 
\underline{98.66}\textsubscript{\textcolor{myGreen}{\footnotesize +0.11}} \\
\bottomrule
\end{tabular}
\label{tab:in-ft}
\vspace{-6pt}
\end{table*}

Our goal here is to evaluate  the value of procedural pretraining with ViTs on
standard image classification tasks.
The experiments are designed to address the following questions.
\begin{enumerate}
    \item Do ViTs acquire information from our abstract non-visual data
    that is useful for vision tasks?
    \item Are the effects of procedural warm-up
    {distinct} and {complementary} to standard large-scale
    visual pretraining?
    \item  What {properties of the procedural data} are responsible for the observed benefits?
    \item How is the learned information {represented within the model's layers?}
\end{enumerate}

\paragraph{Experimental setup.} 
We use a standard ViT-B/16 architecture for experiments with the \textsc{ImageNet-1k} dataset
and a ViT-T/16 for the smaller datasets~\cite{dosovitskiy2020image}.
We expose the models to our procedural warm-up phase as outlined in Section~\ref{sec:methods}
before standard training on natural images.
We use the same hyperparameters throughout the procedural warm-up and the standard training phases, 
with common choices of learning rate, AdamW optimizer, and cosine learning-rate decay (see details in the appendix).

\paragraph{Datasets.}
We use standard datasets that vary in size and complexity
to evaluate performance on image classification:
\textsc{ImageNet-1k}~\cite{deng2009imagenet}, \textsc{Tiny-ImageNet}~\cite{le2015tinyimagenet}, \textsc{Food-101}~\cite{bossard2014food101},  \textsc{CIFAR-10} and \textsc{CIFAR-100}~\cite{krizhevsky2009learning}, and \textsc{STL-10}~\cite{coates2011analysis}.
With \textsc{ImageNet-1k}, we also evaluate the complementarity of procedural warm-up and conventional large-scale visual pretraining.

\paragraph{Baselines.}
Our method is 
meant to provide a generic starting point for training ViTs.
Therefore, we compare it to alternative initialization strategies
also designed to improve subsequent image-based training.
\begin{itemize}[leftmargin=1em]
\item \textbf{Default initialization~\cite{dosovitskiy2020image}:} standard sampling of random weights from a truncated-normal distribution.

\item \textbf{Mimetic initialization~\cite{trockman2023mimetic}:} a structured initialization scheme for attention layers that imitates a diagonal-matrix structure observed in pretrained models.
    
\item \textbf{FractalDB warm-up (sample-matched)~\cite{kataoka2021pretrainingnaturalimages}:}
a form of procedural \emph{visual} data 
also designed to train vision models.
It was originally intended as a sole source of pretraining data
for visual representation learning, rather than a complement like ours.
To match our setting, first,
we adjust and match the number of training examples,
and second, we reset the patch-embedding and classifier layers after pretraining
as described in Section~\ref{sec:pretraining}.
This allows a fair comparison of procedural-visual data (FractalDB) vs.\ procedural-symbolic data (ours).

\end{itemize}
These methods are followed with standard training with identical optimization settings,
such that subsequent differences in performance arise solely from the initialization or warm-up applied to the ViT. 

\subsection{Improving Vision without Images}
\label{sec:4_1}

\paragraph{Background.}
We first evaluate whether a ViT can acquire information that is useful for vision tasks
from our procedural non-visual data.
Recent work already shows that this kind of data can help with 
language, code, and mathematics~\citep{jiang2026proceduralpretraining},
and we now test whether these effects extend to vision.

\paragraph{Setup.}
We apply a procedural warm-up to a ViT with the \textsc{$k$-Dyck} language
($k\!=\!64$, following the setup of \cite{hu2025between}). 
This drives the model to discover a structure of nested dependencies.
The warm-up is run for a fixed budget equivalent to 1\% of the total \textsc{ImageNet-1k} examples used in Sections~\ref{sec:4_1}--\ref{sec:4_2}, after which we proceed with standard training on image datasets as described in Section~\ref{sec:methods}.
This procedural warm-up can be seen as an alternative to standard random
or structured initializations~\cite{huang2020improving,zhao2022zero,trockman2023mimetic}
because its computational cost is minimal and meant to be amortised over many trained models.
We further study the computational budget of the warm-up
in Section~\ref{sec:4_3}.


\paragraph{Results.}
Table~\ref{tab:symbolic-downstream} shows that the procedural warm-up consistently improves downstream performance across all benchmarks. 
On average, procedural warm-up yields a +3.4\% absolute improvement over the default initialization. 
Compared to the Mimetic initialization, the procedural warm-up produces larger improvements, suggesting that our data provides a richer inductive bias than the structured-attention patterns instilled through the Mimetic initialization. 
Our method also outperforms the sample-matched FractalDB, despite the latter containing
\emph{visual} data.
This is a very interesting observation, since the FractalDB data was specifically designed
to capture structures relevant to visual tasks, whereas our data
is much more generic and has no obvious correspondence with visual properties.

Importantly, we observe similar improvements
on \textsc{ImageNet-1k} with a larger ViT-B model, with a
+1.72\%
improvement in top-1 accuracy.
This means that the procedural warm-up
is useful even when the dataset and model size increase.
This suggests that the procedural warm-up and the standard training on natural images
provide training signals that do not completely overlap nor override one another.

\myFrame{\textbf{Take-away.}
ViTs can acquire useful information for vision tasks without images. 
Procedural warm-up on a generic \textsc{$k$-Dyck} language
creates inductive biases beneficial to a broad range of image classification datasets.}
\vspace{3pt}


\subsection{Complementarity with Standard Training} 
\label{sec:4_2}

\paragraph{Background.}
Section~\ref{sec:4_1} shows that procedural warm-up helps prepare ViTs
for one round of training on various target datasets.
We now evaluate whether it helps when the ViTs also undergo
standard pretraining on \textsc{ImageNet-1k},
and whether the benefits persist through subsequent fine-tuning on target datasets.
Prior work with language models~\citep{hu2025between,jiang2026proceduralpretraining}
already shows that procedural data can introduce persistent, complementary structure
that survives pretraining on real language data. 


\paragraph{Setup.}
We first expose the ViT to a small amount of procedural \textsc{$k$-Dyck} data ($k\!=\!64$), using $E_1$ samples (approximately 3.8\,M symbolic examples, corresponding to 1\% of the pretraining budget). The ViT then undergoes standard pretraining on $E_2$ samples from \textsc{ImageNet-1k} (approximately 385\,M image examples). 

We consider two settings for combining procedural warm-up and visual pretraining: \textbf{additive} and \textbf{substitutive}. 
In the \emph{additive} setting, $E_2$ is held fixed and we introduce $E_1$ to test whether procedural warm-up improves the quality of the resulting pretrained model beyond what large-scale visual pretraining alone provides. The resulting models are then fine-tuned on various benchmarks to assess the impact of the warm-up on final downstream performance.

In the \emph{substitutive} setting, we reduce $E_2$ while introducing $E_1$, and measure the reduction $\Delta E_2$ for which training on $E_2\!-\!\Delta E_2$ natural images with $E_1$ procedural examples matches the performance of the $E_2$-only model. 
This quantifies how procedural data can substitute for real images without degrading performance.
All inputs have the same size, so the number of training examples and tokens are proportional to one another.

\begin{figure}[h]
    \vspace{-4pt}
    \centering
    \includegraphics[width=1.0\linewidth]{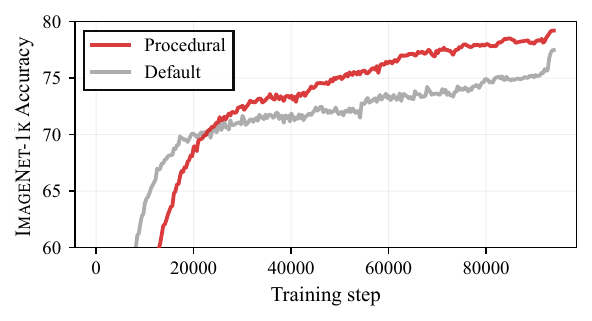}%
    \vspace{-4pt}
    \caption{\textbf{Procedural warm-up leads to a distinct optimization trajectory.}
    With \textsc{ImageNet} pretraining held fixed, a model initialized with a brief procedural warm-up (\textcolor{red}{red}) shows a clearly distinct training curve
    than a default initialization (\textcolor{gray}{gray}).
    This suggests that the procedural warm-up
    provides a qualitatively different training signal
    rather than merely a head-start on standard pretraining.\vspace{-6pt}}
    \label{fig:in-1k-acc}
\end{figure}

\paragraph{Results.}
Procedural warm-up provides a clear and complementary gain on top of large-scale visual pretraining. As shown in Figure~\ref{fig:in-1k-acc}, models initialized with a procedural warm-up follow a distinct optimization trajectory relative to default initialization, and ultimately reach a higher accuracy.

\paragraph{Additive setting.}
With \textsc{ImageNet} pretraining held fixed,
Table~\ref{tab:in-ft} shows that
our procedural warm-up improves performance across benchmarks and outperforms Mimetic/FractalDB. This indicates that the effects \emph{persist after fine-tuning} and \emph{complement standard visual pretraining.}

\begin{figure}[b]
    \vspace{-12pt}
    \centering
    \includegraphics[width=0.6\linewidth]{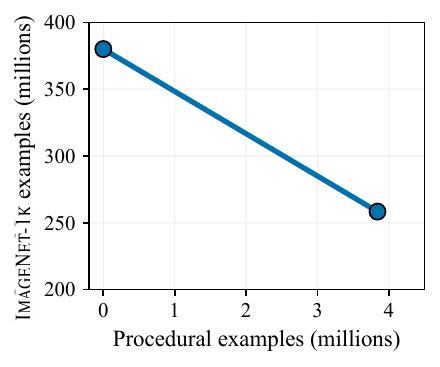}%
    \caption{\textbf{Procedural warm-up reduces the \textsc{ImageNet-1k} data requirements.} 
    Replacing 1\% of the total pretraining budget with 3.8\,M procedural samples allows the model to match the accuracy of a full \textsc{ImageNet-1k} pretraining while using 28\% fewer natural-image samples (about 108\,M fewer images).}
    \label{fig:data_efficiency}
\end{figure}

\vspace{4pt}
\paragraph{Substitutive setting.}
Figure~\ref{fig:data_efficiency} shows that replacing only \emph{1\% of the total pretraining budget} with procedural data yields models that match the performance of those trained solely on \textsc{ImageNet-1k} while using \emph{28\% fewer image examples}.

\myFrame{\textbf{Take-away.} Procedural warm-up provides inductive biases that are \emph{distinct} and \emph{complementary} to standard large-scale visual pretraining, resulting in more efficient training and enhanced performance.}

\vspace{3pt}
\subsection{What are the Important Properties of\\the Procedural Data?}
\label{sec:4_3}

\paragraph{Background.}
The previous sections show that procedural warm-up can speed up the training and improve the performance of ViTs.
We now seek to identify which properties of the procedural data are responsible for these effects.


\paragraph{Setup.}
We start with the \textsc{$k$-Dyck} ($k\!=\!64$) configuration and vary one property at a time.
Specifically, we examine: 
\begin{enumerate}[leftmargin=1.5em]
    \item \textbf{Language type:} we test other data-generating grammars from Table~\ref{tab:formal_languages},
    namely \textsc{WW} and \textsc{$k$-Dyck Shuffle}.
    These alternatives sit at different levels of the Chomsky hierarchy
    and are thus representative of clearly different structural complexity.
    \item \textbf{Order preservation:} we retain the overall distribution of symbols in the generated set of procedural examples, but we shuffle the order of the tokens
    within every sequence
    to break their structural dependencies.
    This assesses whether the gains stem from precise structure or low-order statistics alone.
    \item \textbf{Pretraining length:} we vary the number of gradient steps of the procedural warm-up.
\end{enumerate}

\paragraph{Language type.}
Table~\ref{tab:language-type} shows that the warm-up with both \textsc{Dyck} or \textsc{Dyck Shuffle} yields clear improvements over the default initialization, whereas \textsc{WW} offers no benefit. The strongest gains come from the context-free \textsc{Dyck} grammar, suggesting that a stack-based hierarchy imparts the most transferable inductive bias for vision. The regular \textsc{WW} language, which lacks nested structure, fails to help, indicating that simple repetition does not lead to learning useful mechanisms. The context-sensitive \textsc{Dyck Shuffle} improves performance but lags behind \textsc{Dyck}, implying that excessive structural entanglement weakens transferability.

\paragraph{Order preservation.} 
Shuffling the token order within \textsc{Dyck} sequences eliminates the benefit of procedural warm-up (Table~\ref{tab:dyck-order-shuffled}), reducing performance below the default initialization. 
This confirms that the gains arise from \emph{structural order} rather than simply the token distribution. This shows that the hierarchical dependencies drive the inductive bias, rather than symbol frequency or co-occurrences.

\begin{figure}[t]
\centering
\includegraphics[width=0.59\linewidth]{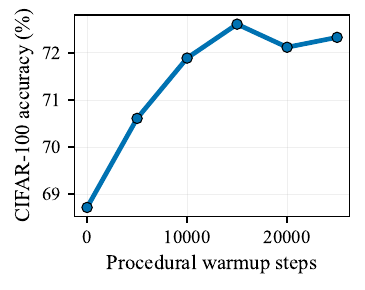}
\vspace{-5pt}
\caption{\textbf{Downstream accuracy as a function of the number of warm-up steps.}
The accuracy peaks at an intermediate value.
With excessive pretraining, the model likely over-specializes to the procedural data,
which hinders adaptation to visual tasks.\vspace{-10pt}}
\label{fig:transferability}
\end{figure}

\paragraph{Procedural pretraining length.} 
As shown in Figure~\ref{fig:transferability}, downstream performance peaks at an intermediate number of procedural warm-up steps, with both shorter and longer runs yielding lower downstream performance. 
This pattern aligns with
prior findings with language models~\citep{springer2025overtrainedlanguagemodelsharder}, where excessive pretraining can hinder fine-tuning adaptability.

\begin{table}[h]
\caption{\textbf{Effect of different formal languages.}
Both \textsc{Dyck} and \textsc{Dyck Shuffle} improve over the baseline, while \textsc{WW} provides no benefit.  
Context-free structure yields the strongest gains, suggesting that hierarchical structured data is key to the observed benefits.}
\centering
\small
\setlength{\tabcolsep}{6.0pt}
\renewcommand{\arraystretch}{1.0}
\vspace{2pt} 
\begin{tabular}{lcc}
\toprule
\textbf{Method} & \textbf{Language type} & \textbf{CIFAR-100} (\%) \\
\midrule
Random initialization & --- & 68.52 \\
\textsc{WW} & Regular & 66.44 \\
\textsc{$k$-Dyck} & Context-free & \textbf{71.98} \\
\textsc{$k$-Dyck Shuffle} & Context-sensitive & 70.11 \\
\bottomrule
\end{tabular}
\label{tab:language-type}
\end{table}

\begin{table}[h]
\caption{\textbf{Effect of removing hierarchical structure in procedural data.}
We shuffle tokens within \textsc{Dyck} sequences to remove hierarchical dependencies
while preserving the token distribution. This degrades the performance, falling below a random initialization.}
\centering
\small
\setlength{\tabcolsep}{4.5pt}
\renewcommand{\arraystretch}{1.0}
\vspace{2pt}
\begin{tabular}{lc}
\toprule
\textbf{Method} & \textbf{CIFAR-100} (\%) \\
\midrule
Random initialization & 68.52 \\
\textsc{$k$-Dyck} & \textbf{71.98} \\
\textsc{$k$-Dyck} (shuffled sequences) & 67.22 \\
\bottomrule
\end{tabular}
\label{tab:dyck-order-shuffled}
\vspace{2pt}
\end{table}

\vspace{-5pt}
\myFrame{\textbf{Take-away.} Procedural warm-up benefits ViTs when the data contains hierarchical structure, when input order is exact, and when training is neither too short nor excessive.}
\vspace{6pt}

\subsection{Which Layers Contain Useful Information\\from Procedural Warmup?}
\label{sec:4_4}

\paragraph{Background.}
The previous sections identified
the effects of procedural warm-up and the responsible properties of the data.
We now shift focus 
to the {model} itself: where and how are the learned inductive biases
expressed within the ViT architecture?
Methods like the Mimetic initialization~\citep{trockman2023mimetic}
act solely on the \emph{attention weights}, while our procedural warm-up
can also affect the MLP layers.


\paragraph{Setup.}
We conduct three analyses to understand how the information learned during the procedural warm-up manifests within the network. 
\begin{itemize}
    \item \textbf{Weight structure.} We test whether the benefit depends on {precise structure} or merely the magnitude of the weights. We shuffle the weight values within a layer, which preserves the distribution of magnitudes while removing any precise structure.
    \item \textbf{Component contributions.}  We test whether the useful information is distributed or located within specific components.
    To do so, we keep the correct weights from the procedural warm-up only for the
    attention or MLP layers, and shuffle the weights within each layer of the other type.
    \item \textbf{Layerwise analysis.} We identify {which layers} contribute most
    by transferring weights from the procedural warm-up
    only for the early, middle, or late 
    $1/3^\textrm{rd}$ of
    transformer blocks and a standard random initialization for the others.
\end{itemize}


\paragraph{Weight structure.}
The model with shuffled weights loses nearly all of the benefits of the procedural warm-up  (Table~\ref{tab:dyck-weights-shuffled}). This confirms that the benefits
are due to the model
solving the pretraining objective and discovering the generating algorithm
from the structure of the procedural data,
rather than simply adjusting the magnitude of initial random weights.

\begin{table}[h!]
\caption{\textbf{Importance of weight structure.} 
Shuffling the weights after procedural warm-up and before image training, eliminates the performance gains,
showing the benefits are due to {precise weight structure}
rather than simply the distribution of their magnitudes.}
\label{tab:dyck-weights-shuffled}
\centering
\small
\setlength{\tabcolsep}{7.5pt}
\renewcommand{\arraystretch}{1.0}
\vspace{2pt}
\begin{tabular}{lc}
\toprule
\textbf{Method} & \textbf{CIFAR-100} (\%) \\
\midrule
Random initialization                  & 68.52 \\
\textsc{$k$~Dyck}                    & \textbf{71.98} \\
\textsc{$k$~Dyck}, all weights shuffled  & 69.51 \\
\bottomrule
\end{tabular}
\end{table}

\paragraph{Component contributions.}
When shuffling weights within attention or MLP layers, the performance degrades substantially (Table~\ref{tab:component-shuffle}).
This means that the learned information is {distributed throughout the network}.
The procedural warm-up establishes coherent relationships across layers
that cannot be replicated with existing schemes like the Mimetic initialization
that only target attention layers~\cite{trockman2023mimetic}.

\paragraph{Layerwise analysis.}
Our selected ablation of early/mid/late layers shows that
the procedural warm-up mostly affects the late layers, which account for most of the gains
over the baseline (Table~\ref{tab:layerwise-transfer}).
This is a surprising observation
because the pretraining of vision models normally relies mostly
on early layers~\cite{lenc2015understanding},
which capture low-level transferable features.
This suggests that the procedural warm-up provides a qualitatively different
type of training signal than standard visual data.


\begin{table}[h]
\vspace{3pt}
\caption{\textbf{Effect of component-wise weight shuffling.} 
Shuffling attention or MLP weights after procedural warm-up reduces downstream performance, showing that the inductive bias is \emph{distributed throughout the network} rather than confined to a single component.}
\label{tab:results}
\centering
\small
\setlength{\tabcolsep}{4.5pt}
\renewcommand{\arraystretch}{1.0}
\vspace{2pt}
\begin{tabular}{lc}
\toprule
\textbf{Method} & \textbf{CIFAR-100} (\%) \\
\midrule
Random initialization                           & 68.52 \\
\textsc{$k$~Dyck}                              & \textbf{71.98} \\
\textsc{$k$~Dyck}, attention weights shuffled & 70.57 \\
\textsc{$k$~Dyck}, MLP weights shuffled       & 70.71 \\
\bottomrule
\label{tab:component-shuffle}
\end{tabular}
\end{table}

\begin{table}[h]

\caption{\textbf{Layerwise transfer.} 
Utilising only subsets of layers from the procedurally pretrained model shows that the
late layers are the most useful. 
Early and middle layers provide limited benefit, suggesting that procedural warm-up imparts an inductive bias expressed in the network's deeper stages.}
\label{tab:layerwise-transfer}
\centering
\small
\setlength{\tabcolsep}{7.5pt}
\renewcommand{\arraystretch}{1.0}
\vspace{2pt}
\begin{tabular}{lc}
\toprule
\textbf{Transferred layers}~~~ & \textbf{CIFAR-100} (\%) \\
\midrule
Random initialization       & 68.52 \\
All Layers    & \textbf{71.98} \\
First 4 layers  & 68.91 \\
Middle 4 layers & 70.19 \\
Final 4 layers   & \underline{71.66} \\
\bottomrule
\end{tabular}
\end{table}

\myFrame{\textbf{Take-away.}
The useful information acquired during the procedural warm-up is encoded in \emph{precise weight structure} of \emph{both attention and MLP} components and concentrated almost exclusively in \emph{late (deep) layers}.}
\vspace{6pt}

\vspace{-2pt}
\section{Discussion}
\label{sec:conclusions}
\vspace{2pt}

We proposed a \emph{procedural warm-up} for ViTs,
a generic strategy to better prepare ViT weights for subsequent learning of vision tasks.
The method trains the model for masked-token prediction
on sequences of abstract tokens generated with formal languages.
These sequences have no semantic meaning or visual structure,
but they force the model to learn simple algorithms, such as a stack mechanism
with sequences of balanced parentheses.
Remarkably, the ViTs trained on this data acquire soft inductive biases
that help the subsequent training on image datasets.
We observe improvements in the speed of convergence and final accuracy.
Our experiments suggest that the procedural warm-up
provides a training signal that is qualitatively different from natural images.
This means that it does not only provide a head-start on standard training, e.g.\ on ImageNet.
Instead, it provides information that would not be acquired otherwise.




\vspace{4pt}
\paragraph{Generality and efficiency.}
Our method is computationally lightweight since it uses
only a small number of training steps and data that can be cheaply generated on the fly.
Even more importantly, the method and the data are extremely generic.
We observe benefits on a range of datasets with no particular tuning,
and prior work~\cite{jiang2026proceduralpretraining,hu2025between,shinnick2025transformers}
even shows benefits on natural language, computer code, and mathematics with a very similar approach.
There is therefore reason to believe that
the procedural warm-up
instils a set of generic or universal computational mechanisms in the model.
This means that a ``warmed-up model''
can be seen as an alternative to a standard random or structured initialization~\cite{huang2020improving,zhao2022zero,trockman2023mimetic}
since the computational cost
can be amortised over many trained models.

\vspace{4pt}
An ambitious prospect is to develop a closed-form instantiation of procedurally-pretrained weights.
Section~\ref{sec:4_3} shows that precise structure in the weights is critical.
Yet the Kolmogorov complexity of the model (i.e.\ its minimal description length) is necessarily small,
since the data-generating algorithm fits in a few lines of code.
Conceptually, it should therefore be possible to summarize a procedurally-pretrained model,
perhaps in a simple formula like in the Mimetic initialization~\cite{trockman2023mimetic} of attention layers.


\vspace{4pt}
Our findings suggest that “learning to see” can be jump-started from procedural data, pointing toward a new path for data-efficient and modality-agnostic pretraining. They also raise broader questions about how inductive biases emerge and propagate within transformers: why do later layers absorb the greatest benefit from procedural warm-up, and how do non-visual priors interact with visual features during fine-tuning? Extending this analysis across architectures, modalities, and other forms of algorithmic data could reveal more general principles underlying transferable and universal representation learning.

\vspace{5pt}

\vspace{4pt}
\paragraph{Limitations.}
(1)~We focus on standard ViTs, but future work could explore other architectures
like Swin~\cite{liu2021swin} and XCiT~\cite{elnouby2021xcit}.
(2)~We experimented with a limited set of procedural generators, with many possible alternatives~\cite{jiang2026proceduralpretraining, wu2022insights}.
(3)~The effects could be evaluated in other settings like segmentation, OOD generalization, and multimodal models.


{
    \small
    \bibliographystyle{ieeenat_fullname}
    \bibliography{main}
}
\clearpage
\normalsize     
\onecolumn

\appendix
\renewcommand{\thesection}{\Alph{section}}
\renewcommand{\thesubsection}{\Alph{section}.\arabic{subsection}}


\addcontentsline{toc}{section}{Appendix}

\addcontentsline{toc}{section}{Appendix}

\section{Procedural Warm-Up Details}

We pretrain the ViT-T/16 configuration used throughout the paper for 15k steps using masked-token prediction. 
Table~\ref{tab:train-hparams} summarizes the training hyperparameters.

\setlength{\tabcolsep}{4pt}

\begin{table}[h]
\centering
\footnotesize
\begin{tabular}{r l}
\toprule
\textbf{Setting} & \textbf{Value} \\
\midrule
Batch size & 256 \\
Training steps & 15{,}000 \\
Mask ratio & 0.5 (close-only) \\
Optimizer & AdamW \\
Learning rate & 2e$^{-3}$ \\
Weight decay & 0.05 \\
Betas & (0.9, 0.999) \\
LR schedule & Cosine decay \\
Warmup steps & 1{,}000 \\
\bottomrule
\end{tabular}
\caption{Training hyperparameters for the procedural warm-up stage.}
\label{tab:train-hparams}
\end{table}

For sampling from the \textsc{$k$-Dyck} and \textsc{$k$-Dyck Shuffle} languages, we draw opening tokens from $k_{\text{open}}$ types and closing tokens from $k_{\text{close}}$ types, using an opening probability of $p_{\text{open}} = 0.6$ whenever opening is structurally permissible.

In Section~\ref{sec:4_3}, when varying the language type, we modify only the underlying symbolic generator. 
When varying the pretraining length, we train a separate model for each duration by adjusting the number of training steps accordingly.

\section{Image-Based Training Details}

For the experiments in Section~\ref{sec:4_1} (Table~\ref{tab:symbolic-downstream}), we follow the standard ViT training setup used in~\cite{xu2023initializing}, including AdamW optimization, cosine learning-rate decay, linear warmup, RandAugment, Mixup, CutMix, and label smoothing. 
Across all datasets, we use a base learning rate of 2e$^{-3}$ and train for 300 epochs with 50 warmup epochs. 
We use a batch size of 512 for all models, except for ViT-B on \textsc{ImageNet-1K}, where we use a batch size of 4096.

For the experiments in Section~\ref{sec:4_2} (Table~\ref{tab:in-ft}), we further fine-tune the \textsc{ImageNet-1K} ViT-B checkpoints on the smaller datasets.
We use the same training setup as above (with batch size 512), but train for 50 epochs with 5 warmup epochs and a base learning rate of 5e$^{-4}$.

\section{Additional Results}
\begin{table}[h]
\vspace{-7pt}
\centering
\footnotesize
\setlength{\tabcolsep}{7.0pt}
\renewcommand{\arraystretch}{0.95}
\vspace{1pt}
\begin{tabular}{lccccc}
\toprule
\phantom{\raisebox{.15ex}{$\blacktriangleright$}}
\textbf{Method} &
\textbf{\textsc{Tiny-ImageNet}} &
\textbf{\textsc{Food-101}} &
\textbf{\textsc{CIFAR-10}} &
\textbf{\textsc{CIFAR-100}} &
\textbf{\textsc{STL-10}} \\
\midrule
\phantom{\raisebox{.15ex}{$\blacktriangleright$}}
Default init.
& 55.42 & 74.52 & 91.29 & 68.52 & 60.52 \\
\phantom{\raisebox{.15ex}{$\blacktriangleright$}}
FractalDB warm-up
& 55.17 & 74.25 & 88.98 & 64.61 & 58.62 \\
\raisebox{.15ex}{$\blacktriangleright$} FractalDB warm-up + emb.
& 55.64 & 75.99 & 90.44 & 67.35 & 65.62 \\
\phantom{\raisebox{.15ex}{$\blacktriangleright$}}
\textbf{Procedural warm-up (ours)}
& \textbf{58.20} & \textbf{79.47} & \textbf{92.81} & \textbf{71.98} & \textbf{66.48} \\
\bottomrule
\end{tabular}
\caption{Comparison with warm-up on FractalDB when retaining pretrained embeddings. Unlike the main setting where embeddings are reset for consistency across methods, this variant preserves FractalDB’s patch embeddings. Our method still outperforms this baseline.}
\label{tab:fractaldb-embeddings}
\vspace{-7pt}
\end{table}

\begin{table}[h]
\vspace{-4pt}
\centering
\footnotesize
\setlength{\tabcolsep}{5.0pt}
\renewcommand{\arraystretch}{0.9}
\begin{tabular}{lcccc}
\toprule
& $k$=$16$ & $k$=$32$ & $k$=$64$ & $k$=$80$ \\
\midrule
Procedural warm-up
& 69.05 & 69.78 & \textbf{71.98} & 71.83 \\
\bottomrule
\end{tabular}
\caption{CIFAR-100 top-1 accuracy (\%) 
with varying vocabulary size ($k$).
All settings outperform random initialization, with peak performance at an intermediate vocabulary size.
}
\vspace{-4pt}
\end{table}

\begin{table}[h]
\vspace{-6pt}
\centering
\footnotesize
\setlength{\tabcolsep}{7.0pt}
\renewcommand{\arraystretch}{0.95}
\vspace{1pt}
\begin{tabular}{lcc}
\toprule
\phantom{\raisebox{.15ex}{$\blacktriangleright$}}
\textbf{Method} &
\textbf{\textsc{Tiny-ImageNet}} &
\textbf{\textsc{C100}} \\
\midrule
\phantom{\raisebox{.15ex}{$\blacktriangleright$}}
Default init.
& 55.42 $\pm$ 0.66 & 68.52 $\pm$ 0.27 \\
\phantom{\raisebox{.15ex}{$\blacktriangleright$}}
Mimetic init.
& 57.20 $\pm$ 0.62 & 70.72 $\pm$ 0.39 \\
\phantom{\raisebox{.15ex}{$\blacktriangleright$}}
FractalDB warm-up
& 55.17 $\pm$ 0.33 & 64.61 $\pm$ 0.51 \\
\phantom{\raisebox{.15ex}{$\blacktriangleright$}}
\textbf{Procedural warm-up (ours)}
& \textbf{58.20 $\pm$ 0.22} & \textbf{71.98 $\pm$ 0.74} \\
\bottomrule
\end{tabular}
\caption{Mean $\pm$ standard deviation over 3 random seeds on representative datasets of differing scale, showing consistent gains.}
\label{tab:variance}
\vspace{-8pt}
\end{table}


\end{document}